\pdfoutput=1
\documentclass[11pt]{article}

\usepackage[final]{coling}

\usepackage{times}
\usepackage{latexsym}

\usepackage[T1]{fontenc}
\usepackage[utf8]{inputenc}

\usepackage{microtype}

\usepackage{inconsolata}
\usepackage{booktabs}
\usepackage{graphicx}

\title{Personalized Large Language Model Assistant with Evolving Conditional Memory}

\author{Ruifeng Yuan, Shichao Sun, Yongqi Li, Zili Wang, Ziqiang Cao, Wenjie Li \\
  The Hong Kong Polytechnic University \\
  INF Technology (Shanghai) co., Ltd. \\
  Institute of Artificial Intelligence, Soochow University, China \\}
\begin{document}
\maketitle
\begin{abstract}
With the rapid development of large language models, AI assistants like ChatGPT have become increasingly integrated into people's works and lives but are limited in personalized services. In this paper, we present a plug-and-play framework that could facilitate personalized large language model assistants with evolving conditional memory. The personalized assistant focuses on intelligently preserving the knowledge and experience from the history dialogue with the user, which can be applied to future tailored responses that better align with the user's preferences. Generally, the assistant generates a set of records from the dialogue dialogue, stores them in a memory bank, and retrieves related memory to improve the quality of the response. For the crucial memory design, we explore different ways of constructing the memory and propose a new memorizing mechanism named conditional memory. We also investigate the retrieval and usage of memory in the generation process. We build the first benchmark to evaluate personalized assistants' ability from three aspects. The experimental results illustrate the effectiveness of our method.

\end{abstract}

\section{Introduction}

The large language model (LLM) has developed rapidly recently. A LLM assistant based on dialogue system such as ChatGPT has been used by millions of people and receives great interest from the public. These systems can chit-chat with the human user and provide assistance by answering questions or executing instructions. 

\begin{figure}[t]
\centering
\includegraphics[width=0.9\columnwidth]{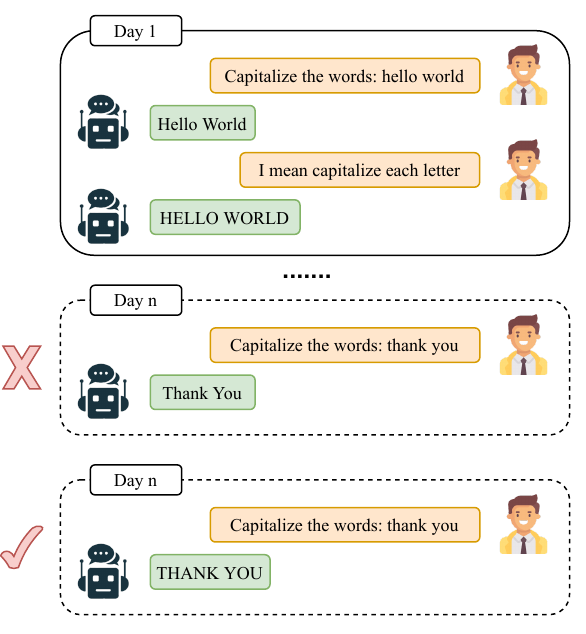} 
\caption{An example of what a personalized LLM assistant with evolving memory can do.}
\label{fig1}
\end{figure}

Unfortunately, one main problem for the current AI assistant is that the information in one dialogue only lasts for one dialogue session. When a user starts a new session, the AI assistant is reset to its initial state and will not preserve any information from previous dialogue sessions. This prevents the AI assistant from acquiring knowledge and experience from the long-term dialogue, to improve the quality of its responses, like humans do. 
As shown in the example in Figure \ref{fig1}, when the user asks the assistant to help him capitalize some words, the assistant chooses to capitalize the first letter of each word, which is not what the user expected. Then the user provides more detailed feedback to avoid ambiguity and get the answer he wants. If the assistant is a human, he can learn that ``capitalize" refers to capitalizing each letter according to this user's preference and can give the more suitable response in the later dialogue, while a LLM assistant fails to achieve this. Therefore, our aim is to explore how to endow the LLM assistant with such an essential capability that allows it to learn knowledge and experience from previous dialogues and apply them to later ones.

%Unfortunately, one main problem for the current state-of-the-art systems is that the conversation only last for one single dialogue session. When a user starts a new chatting session, the assistant will not preserve information from any previous dialogue sessions. In this case, such major aspect missing from these system inevitably brings inconvenience to users. Considering it is a simple fact that human conversations can take place over long time frames, failing to continue a conversation happened few hours ago with the assistant is frustrating. More importantly, every time a new dialogue session begins, the AI assistant is reset to its initial state. This prevents the AI assistant from acquiring knowledge and experience from long-term dialogue, like humans do, to improve the quality of its responses. As shown in the example in Figure \ref{fig1}, when user ask the assistant to help him to capitalize certain words, the assistant choose to capitalize the first letter of each word, which is not what the user expected. Then the user provides more detailed feedback to avoid ambiguity and get the answer he wants. If the assistant is a human, he can learn that "capitalize" referring to capitalize each letter for this user and can give the right respond in the later conversation, while a LLM assistant fails to achieve this. Therefore, our aim is to explore how to endow the LLM assistant with such an essential capability that allows it to learn knowledge and experience from previous dialogues and apply them to later conversations.

\begin{figure}[t]
\centering
\includegraphics[width=1\columnwidth]{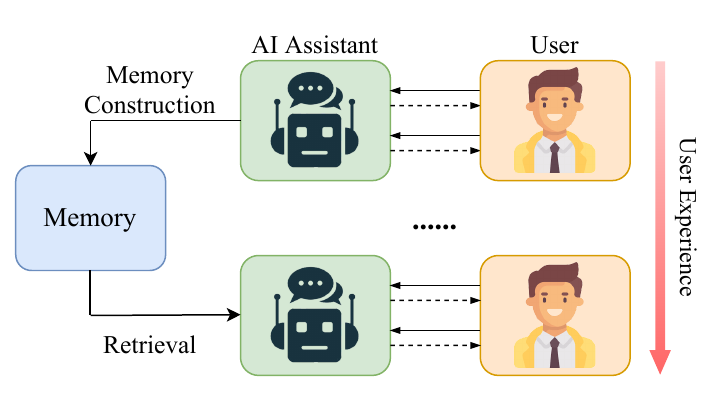} 
\caption{The framework of personalized assistant with evolving memory.}
\label{fig2}
\end{figure}

In this work, we propose to adopt a memory-based framework as a wrapper outside the LLM assistant to address this challenge. Such personalized LLM assistant mainly consists of three parts: an existing LLM assistant, an evolving memory, and a prompt-based framework conducting interactions between the assistant and the memory. As shown in Figure \ref{fig2}, the framework utilizes the ongoing dialogue and LLM assistant itself to construct memory records and store them in the memory. Every time the assistant generates a response, we use a retrieval model to retrieve a series of memory records related to the user input from the memory and combine the information with the user input to generate a better response. Using a memory-based framework has a few advantages. This method does not require additional finetune of the LLM, which is crucial when most of the current LLMs are unattainable or expensive to finetune. It also allows the framework to be easily deployed on any device with an LLM API, without the need for a GPU. Moreover, the memory file is the only place that stores user-specific information, so it is private and can be easily transferred between users' own devices.

To further explore the details of the memory-based personalized LLM assistant, we conduct our research from three aspects. 1) The memorizing mechanism is a key point for a memory-based model. The way to convert a large amount of dialogue history into effective memory records is crucial but challenging for the performance of the entire model. Two straightforward ideas based on history and summaries from previous works \cite{xu2021beyond, park2023generative} are tested under this setting. By combining the advantages of both intuitive ideas, we propose conditional memory, where we believe that the memorization timing and the content of the memory records should be conditional. This allows the memory to capture detailed information while being relatively concise. 2) We also present a self-reflection mechanism for memory retrieval, which allows the LLM assistant to determine whether the retrieved information is sufficient to respond and whether to conduct further searches. 3) Considering there is no existing dataset that is appropriate for testing the ability of an LLM assistant with long-term memory, we build three datasets using GPT-4 with different ability aspects: continuing previous dialogue, learning new knowledge, and learning from human feedback. 

Our contribution can be concluded as follows:
%(1) We propose a memory-based dialogue system as a solution for an evolving LLM assistant that continually improves itself through daily usage.
(1) We propose a plug-and-play framework to facilitate the personalized LLM assistant with external evolving memory. 
%(2) We investigate the framework of the evolving LLM assistant and propose a new memorizing mechanism.
(2) For the crucial memory design in personalized LLM assistants, we propose an effective conditional memory mechanism and a self-reflection memory retrieval strategy.
%(3) We construct three test datasets on different aspects to test the model's abilities in learning from the dialogue history.
(3) We construct the first benchmark to evaluate the personalized assistants' ability to capture user preferences for tailored responses.

\section{Related Work}

%\subsection{Retrieval-Based Dialogue System}
%Retrieval-based dialogue system
Retrieval-based dialogue system is a popular topic in the field and has long been studied as an important paradigm. Open-domain chit-chat dialogue system is a typical embodiment of this paradigm. Such dialogue systems are challenging due to the diversity of possible responses to a single dialogue history. Hence, external knowledge is necessary for avoiding safe but boring responses. These dialogue systems usually retrieve information from the dialogue corpora and focus on how to integrate the retrieved dialogue into the generation process \cite{wu2019response,cai2018skeleton,yang2019hybrid,gupta2020controlling,paranjape2021hindsight}. Other work further extends the external information to different types of knowledge such as knowledge bases and external documents \cite{dinan2018wizard,lian2019learning,qin2019conversing,wu2021controllable,komeili2021internet}.

%conversational qa
Another type of work using the paradigm is conversational question answering. Considering it is a combination of both dialogue and QA, a retrieval system that searches the answer information from the knowledge base or document corpus is inevitable for improving the accuracy of the answer. Different from the open-domain dialogue system, conversational QA pays more attention to the accuracy of the retrieval including the retrieval optimization and query understanding \cite{qu2019bert,christmann2019look,qu2020open,vakulenko2021question,adlakha2022topiocqa}.

%大模型
Recently, as a stronger version of the dialogue system, LLM-based chatbots such as ChatGPT can also be augmented by retrieval. These systems are capable of providing a more informative and faithful answer by retrieving question-related web documents directly from the Internet \cite{he2022rethinking,ram2023context,liu2023reta}. There also exist many toolkits for building this type of system such as LangChain. 

\begin{figure*}[t]
\centering
\includegraphics[width=2\columnwidth]{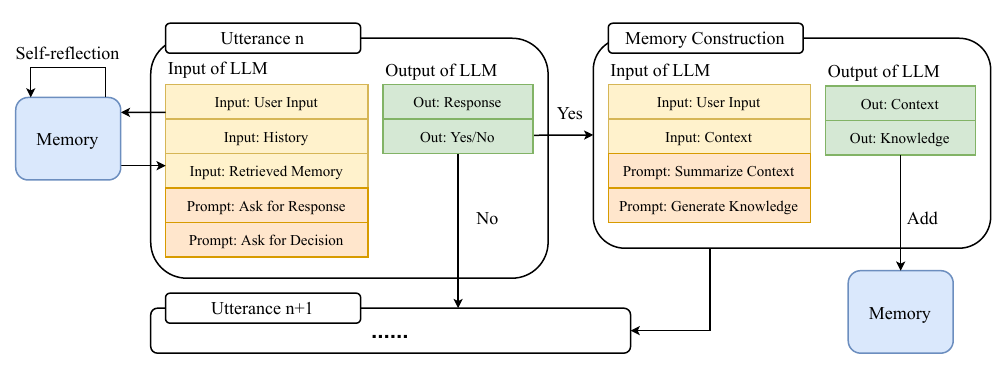} % Reduce the figure size so that it is slightly narrower than the column. Don't use precise values for figure width.This setup will avoid overfull boxes.
\caption{The LLM assistant using conditional memory with self-reflection retrieval.}
\label{fig3}
\end{figure*}

\begin{table*}[t]
\centering
%\resizebox{.95\columnwidth}{!}{
\begin{tabular}{lccp{10cm}}
    \hline
    Memory & Frequency & Length & Example \\
    \hline
    History & 1 & 52.6 & Assistant: Do you prefer going to the gym? User: I like the controlled environment. I don't want to depend on the weather.\\
    \hline
    Summary & 0.2 & 74.6 & The user and the assistant discussed their preferences for running and going to the gym. ... They also discussed the availability of parks and running trails in the area. \\
    \hline
    Conditional & 0.5 & 46.8 & Context: The user and the assistant are discussing the inconvenience of running outdoors due to weather conditions. Knowledge: User likes gym workouts due to a controlled environment and not having to depend on weather. \\

    \hline
\end{tabular}
\caption{Statistics and examples of the three types of memory generated from chit-chat dialogue in multi-session chat dataset \cite{xu2021beyond}.  ``Frequency" refers to the average number of memory records generated per utterance. ``Length" represents the average length of the generated memory records.}
\label{tablecase}
\end{table*}
%``LLM Call" refers to the average LLM api call per utterance (response and memory construction) for the LLM assistant
%Summary: The user and the assistant discussed their preferences for running and going to the gym. They both prefer the controlled environment of a gym due to unpredictable weather. The user suggested getting a treadmill for home use, but the assistant is considering joining a local gym for more workout options. They also discussed the availability of parks and running trails in the area and the possibility of joining or starting a running club.

%Context: The user and the assistant are discussing the inconvenience of running outdoors due to weather conditions, and the user suggests considering a treadmill for at-home use; Knowledge: User prefers gyms with controlled environments, and suggests buying a treadmill for running at home to avoid rainy weather.

%区别主要是搜的东西不一样，不需要构造
%The main idea of these retrieval-based dialogue system is that they need to retrieve dialogue-related information from existing knowledge or data sources through the retrieval model, and use it to enhance the diversity or factual accuracy of the respond. In this case, these models mainly focused on improving the accuracy of the retrieval and the usage of external information in respond generation. 

%memory-based

Memory-based models can be considered a special type of retrieval-based model. Instead of retrieving information from an existing source, the memory-based models retrieve useful information from its memory while keeping add more memory records to it. This also makes how to construct the memory a key point in the research of this type of model. Some researchers \cite{park2023generative} apply the memory-based dialogue system on interactive agents where some AI agents simulate humans in a virtual environment. A history-based memory system is used to help the agents achieve long-term interaction. Other researchers \cite{xu2021beyond} use it on continuing a multi-session dialogue and propose a summary-based memory system. In our work, we also adopt a solution based on a memory system and target its application in building an personalized LLM assistant.

\section{Method}
The memory-based framework of the personalized LLM assistant consists of two parts: the memory construction during the dialogue and the retrieval and use of the memory records in response generation. In this section, we mainly focus on the methods and more details such as prompt design are shown in the Appendix.

%In figure \ref{fig2}, we display a brief framework about how the memory-based evolving LLM assistant works. After finish each dialogue session, we use the LLM assistant to build memory records together with the generation of the respond. After that, the records are stored in a memory bank. Concerning the usage of the memory, each time there comes a user input, a retrieval model is used to pull related memory records from the memory bank associated with the user input. This information, along with the user's input, then works to generate a more refined response.

\subsection{Memory Construction}

%Memory construction is a key problem in a memory-based model. Different forms or origins of memory can have a important impact on the effectiveness of the model. In our work, we investigate three different types of memory and compare their influence on the model. 

The construction of memory is a crucial problem in a memory-based model. The varying forms or sources of memory can significantly affect the efficacy of the model. In our research, we investigate three distinct memory types and compare their respective influence on the assistance's performance. The example and statistic of the three types of memory is shown in Table \ref{tablecase}.

\paragraph{History-Based Memory} Utilizing the unprocessed version of the dialogue history directly as the memory is a simple but practical approach, especially considering it does not require any additional generation from the LLM assistant. A similar approach has been widely adopted in question-answering or knowledge-grounded dialogue. Here, we concatenate both the role of the dialogue (user or assistant) and its content with a colon as the memory record. All dialogue history is added into the memory sequentially.

\paragraph{Summary-Based Memory} Dialogue summaries can also be used as memory. As stated by \cite{xu2021beyond}, using dialogue summaries as memory provides more context information than using dialogue history for the retrieval and generation process. In this work, we use GPT-4 to generate a summary for each dialogue session. Here, a dialogue session stands for multiple utterances that are continuous in time and content. These generated summaries are stored as memory records.

%However, those approaches have two potential drawbacks: (i) there is a lot of context to store, and hence retrieve from; (ii) no processing has been done on that content, so the reading, retrieving and combining operations required to generate an answer leave a lot of work for the model to do.

\paragraph{Conditional Memory} Although the above approaches are commonly used, they have two potential drawbacks: (1) They do not distinguish what information is important and worth memorizing in detail. In terms of history-based memory, it treats all information equally, resulting in a large amount of redundancy in the memory. In terms of summary-based memory, they store important information in a simplified way, often lacking detail for key contents. In fact, not all information is worth preserving, and it is important to know when to memorize. (2) The effectiveness of memory records may be conditional, limited only to specific situations. Therefore, the background context information is crucial for a memory record during memory retrieval. History-based memory fails to store any context information, while summary-based memory can only provide the background of the whole dialogue rather than specific context. Therefore, we believe that the timing and content of memory are both conditional.

%under certain conditions
Memorizing only the important information is crucial to reduce redundancy in the memory and highlight key points. We believe that the importance of an utterance is related to the amount of information it contains that is unknown to the LLM. For example, unimportant utterances stand for simple instructions or questions and phatic communication like greetings. These utterances usually occupy a large proportion of the dialogue. For important utterances, it could contain domain knowledge unfamiliar to LLM, user's requirement, or personal preference. In this work, we combine the instruction prompt and the good/bad examples with GPT-4 to determine whether each utterance is important enough to be stored in memory. As shown in Figure \ref{fig3}, the decision is made together with the response generation in one LLM call using some prompt engineering techniques. Hence, it does not largely increase the cost of LLM.

%Selective memory of important information is an effective method to reduce redundancy in memory bank and to highlight key points. We believe that the importance of a utterance is related to the amount of information it contains that the LLM does not know. For example, unimportant utterance stands for simple instructions or questions and phatic communication like greeting. The former one only contains information that LLM knows while the latter just do not contain any effective information. However, these content usually occupies a large proportion of the dialogue. As for the important utterance, it can contain domain knowledge that is unfamiliar for LLM, user's requirement or personal preference. In our framework, we combine the prompt and the good/bad in-context example with GPT-4 make the judgement each utterance about whether it is important enough to be stored in the memory. As shown in the Figure \ref{fig3}, it is worth to notice that the judgement is made together with the respond generation in one LLM call using some prompt engineering techniques. Hence, the selection of the memory do not largely increase the cost of framework.

Based on our assumption that dialogue information is conditional, we propose to divide one memory record into two parts: context and knowledge. Context is a summary that describes the background of the target utterance based on its neighbors, and knowledge is the high-level conclusion about what the assistant learns from the target utterance. By setting different objectives for the two parts, we can achieve a more comprehensive memory record that takes into account both retrieval and generation abilities. We use the GPT-4 to generate the two parts with different prompts. 

\paragraph{Multi-View Memory} There is no limitation that only one type of memory record is allowed in the memory. It is possible that a single type of memory record is not sufficient to contain all information related to user input. Here, we also try to find out whether the multi-view memory provides more diverse information for the response generation.

\subsection{Memory Retrieval and Application}

\paragraph{Memory Retrieval} We use dense retrieval to search the related memory records based on the user input. We encode the user input and each memory record to a vector using SimCSE model \cite{gao2021simcse}. Considering the scale of the memory is not large enough to use the FAISS, we calculate the cosine similarity between the representation of the memory record and the user input and use the top $K$ records for response generation. 

\paragraph{Self-Reflection on Memory} To utilize the memory in response generation, a common practice is to concatenate all the retrieved memory records with the original prompt as background information. However, we can not ensure that retrieved memory records are all useful, and there might even be situations where there is no relevant information at all, which can be misleading for the LLM assistant. Here, we propose to adopt a self-reflection mechanism to address this problem. This self-reflection process contains two steps. After we obtain the retrieved memory records, we ask the LLM to make a decision about whether the retrieved information is enough for it to respond to the user input. If the answer is yes, the LLM assistant selects the related information from these memory records as output. If the answer is no, the LLM assistant is required to improve the original query (user input) to find more information. In practice, the improvement is reified as generating some keywords/phrases representing the missing information. We can extend the original query with these keywords for another retrieval. The self-reflection process can be conducted multiple times until the LLM assistant finds enough information.

\section{Dataset}

\begin{figure*}[t]
\centering
\includegraphics[width=2\columnwidth]{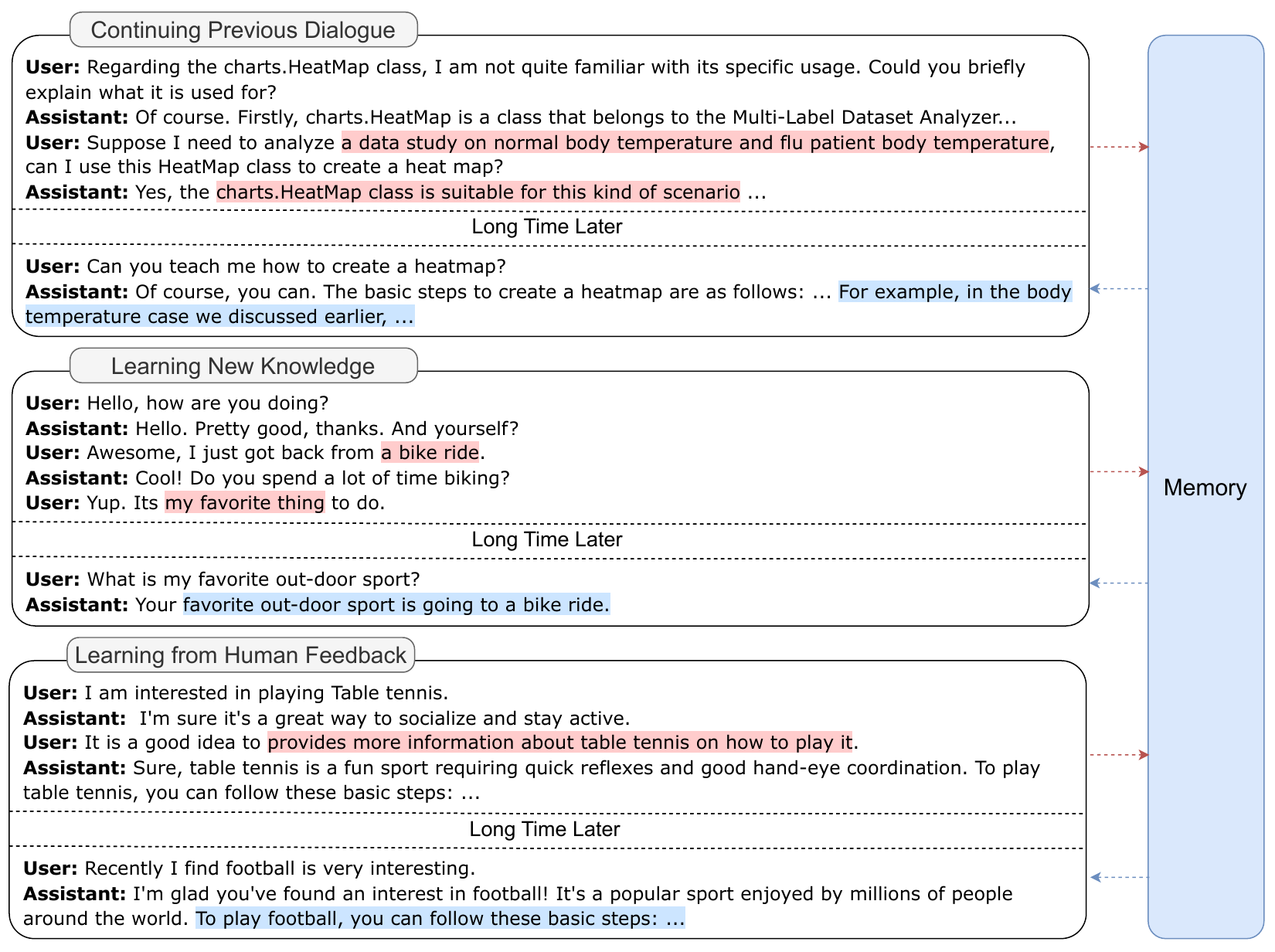} % Reduce the figure size so that it is slightly narrower than the column. Don't use precise values for figure width.This setup will avoid overfull boxes.
\caption{The examples of the three constructed test datasets.}
\label{fig4}
\end{figure*}

\begin{table}[t]
\centering
\begin{tabular}{lcccc}
    \hline
     & Sample & Session & Turn & Turn\\
     Dataset  &  Num &  Source & Source & Test\\
    \hline
    Knowledge & 300 &  4   & 23.5   &  3.8  \\
    Feedback  & 200 &  1   & 5    &  1   \\
    Continue  & 200 &  1   & 4    &  1    \\
    \hline
\end{tabular}
\caption{The statistics of datasets. ``Session Source" and ``Turn Source" refers to the average session number and utterance number of source dialogues.``Turn Test" represents the average utterance number for testing.}
\label{table1}
\end{table}

To evaluate a personalized assistant, we construct datasets from following aspects: continuing previous dialogue, learning new knowledge, and learning from user feedback. Data samples from the three datasets follow the same structure, which contains a dialogue history as the source for constructing memory and some testing dialogue utterance with reference responses. All the three datasets are synthetic data based on GPT-4. In this case, due to the high cost of GPT-4 usage, the datasets are only small-scale datasets. The statistics of the datasets are shown in Table \ref{table1}. Here, we briefly introduce the three datasets and their construction process. The human evaluation for datasets and construction details are in the appendix. This allows the researchers to generate more data samples with GPT-4.

\subsection{Continuing Previous Dialogue}

For a personalized assistant with memory, continuing a previous dialogue session is a basic ability. It requires the model to use information from previous dialogues to enhance the diversity or readability of the next response. For example, in Figure \ref{fig4}, when a question related to heatmap is asked by the user, the assistant can utilize the project discussed under the topic of heatmap before as an example in its response. To construct such a dataset, we convert it to a dialogue continue writing problem. Given the dialogue between the user and the AI assistant, we use the GPT-4 to continue writing a new utterance that contains information from the existing context. This generated utterance is used for testing and the existing dialogue is used as the history for memory construction. In practice, we adopt a code discussion dialogue between AI assistant and user \cite{yang2023refgpt} as source dialogue, since it has rich contextual connections.

%For a dialogue system that contains memory, we think that using the information from the previous dialogue to enhance the diversity or readability of the respond is a basic ability for it. Such ability also allows the assistant to continue a dialogue from a previous dialogue session. For example, in Figure \ref{fig4}, when a question related to heatmap is asked by the user, the assistant can utilize the project has discussed under the topic of heatmap before as an example in its respond. In this case, it provide a more personal and detailed experience for the user. To build up such dataset, we convert it to a dialogue continue writing problem. We first find a set of dialogue between the user and the AI assistant, and use the GPT-4 to generate an utterance that closely follows the existing dialogue The utterance is required to contain information from the context. This generated utterance is used for testing and the existing dialogue is used as the history for memory construction. In practice, we adopt the code discussion dialogue between AI assistant and human from \cite{yang2023refgpt}.

\subsection{Learning New Knowledge}

Another important ability of the personalized assistant is acquiring personal knowledge from its dialogue with the user. We construct such data based on a multi-session chat (MSC) dataset \cite{xu2021beyond}. The MSC dataset contains a multi-session chit-chat between two human and their personal settings. And these personal settings can be inferred from the dialogue. In our dataset, as shown in Figure \ref{fig4}, we assume that one of the two participants in the dialogue is the user and the other one is the assistant. These dialogue sessions are considered dialogue history. Then we use GPT-4 to convert the personal settings of the user to question-answer utterances aiming to test whether the assistant learns these knowledge in the dialogue.

%The construction process of MSC is to ask two human workers to have conversation based on these personal settings. 

\subsection{Learning from Human Feedback}

The last ability we test is whether the personalized assistant can learn from the user's feedback or preference and use it properly in a similar situation. For instance, in Figure \ref{fig4}, the user seeks for information about table tennis and provides feedback about requiring more information on how to play it. In the future, when the user asks for information about another sport football, the assistant can directly add information about how to play it in its response. 
In terms of dataset construction, given a seed instruction/question sampled from the seed set of self-instruction \cite{wang2022self}, we first generate a response using GPT-4 and generate suggestions to improve the response. These suggestions are regarded as pseudo human feedback. Then we use the suggestions to improve the response. In this case, we obtain a two-turn dialogue with human feedback. We also add a background dialogue before it with GPT-4 using a similar method with \cite{yang2023refgpt}. The concatenation of the two dialogues is used as dialogue history. For testing, we generate a similar instruction or question with the seed one and generate a reference response that also follows suggestions given in dialogue history. The assistant is expected to learn from previous feedback and directly generate the reference response in the later dialogue. 

\section{Experiment}

\subsection{Application Details}

Considering we aim to test the effectiveness of different memory mechanisms without the limitation of the LLM, we use GPT-4, current strongest LLM, as the backbones of LLM assistants in all experiment. For the evaluation metrics, we use an N-gram similarity metric, Rouge \cite{lin2004rouge} including Rouge-1 and Rouge-2. We also use the GPT-4 for more complicated evaluations, which are shown in the next subsection. For the hyperparameter, we set the number of the retrieved memory records $K$ as 3 for history-based memory and conditional memory, and we set $K$ as 1 for summary-based memory. It is also worth to notice some source dialogue history can be relatively short. To simulate the long-term use of an assistant, we merge the memory of every 10 data samples into one in the experiment of learning from feedback and continuing previous dialogue.

%As for the LLM assistant and dataset construction, we use the default generation hyperparameters. As for the evaluation, we set temperature to 1 to obtain a definitive answer.

\begin{table*}[t]
\centering
\begin{tabular}{lcc}
    \hline
     Dataset  &  Cond+His vs Cond &  Cond+Sum vs Cond  \\
    \hline
    Learning New Knowledge & 0.58\ /\ \textbf{0.63} &   \textbf{0.69}\ /\ 0.63     \\
    Learning from Feedback   & 27\%\ /\ 29\%\ /\ \textbf{44\%}  &   \textbf{34\%}\ /\ 36\%\ /\ 30\%          \\
    Continuing Previous Dialogue   & \textbf{28\%}\ /\ 50\%\ /\ 22\%  &   \textbf{29\%}\ /\ 53\%\ /\ 18\%    \\
    \hline
\end{tabular}
\caption{The evaluation result of combining conditional memory (Cond) with history (His) and summary (Sum). For evaluation metrics, we adopt GPT-score for learning new knowledge, and GSB test for the other two datasets.}
\label{table4}
\end{table*}

\subsection{Automated Evaluation with GPT-4}

There are three different types of GPT-4 evaluation in our experiment: scoring, comparing, and multiple choice. To ensure the fairness in GPT-4 evaluation \cite{wang2023large}, we list the details and related prompts in the appendix. 

Scoring \cite{liu2023gpteval, fu2023gptscore} focuses on giving a predicted response a score between a predefined margin when the reference answer is clear. We call it GPT-score. The GPT-4 is required to give a 0-2 score for a response. Here, ``0” represents the response is totally wrong, ``1” represents the response contains part of information in the reference answer and ``2” represents the response contains all information. 

Comparison \cite{zheng2023judging} aims to conduct a Good/Same/Bad test (GSB) between two candidates, which is applied in a more complicated evaluation situation. The GPT-4 is used to make a decision about which candidate is better given the reference response and other necessary information. Multiple choice is similar to comparison, but the GPT-4 needs to choose the best candidate among multiple options.

We use GPT-score for learning new knowledge. As for the GSB test, it is used for learning from feedback and continuing previous dialogue. The main reason for such difference is that learning new knowledge provides clear reference answers where a direct comparison is more precise, while the other two tasks have more open answers that are more suitable for the GSB test.

%There are three different types of GPT-4 evaluation in our experiment: scoring, comparing, and multiple choice. The detailed prompt for evaluation is listed in the appendix. 

%Scoring focuses on giving a predicted response a score between a predefined margin when the reference answer is clear. We call it GPT-score and use it to evaluate the model ability in learning new knowledge. The GPT-4 is required to give a 0-2 score for a response. Here, ``0” represents the response is totally wrong, ``1” represents the response contains part of personal information in the reference and ``2” represents the response contains all information. 

%Comparison aims to conduct a Good/Same/Bad test (GSB) between two candidates, which is applied in a more complicated evaluation situation. The GPT-4 is used to make a decision about which candidate is better given the reference response and other necessary information. Note that the GPT-4 has a position bias and candidates at a specific position (first or second) may have more advantages. Hence, we conduct the evaluation twice and exchange the position of the two candidates. Only when the two evaluation meets the result of each other, the result is definite. If the results are different, we consider it as a tie. 

%Multiple choice is similar to comparison, but the GPT-4 needs to choose the best candidate among multiple options. For the position bias problem, we solve it by randomly placing all candidates in different positions. 

\subsection{Experiment Results}

\begin{table}[t]
\centering
\begin{tabular}{lccc}
    \hline
     Memory  &  GPT-score &  Rouge-1 & Rouge-2 \\
    \hline
    No memory & 0.15 &   33.17   &  11.46   \\
    History   & 0.56 &   40.42   &  15.49   \\
    Summary   & 0.57 &   43.92   &  16.63    \\
    Conditional   & \textbf{0.63} &   \textbf{46.08}   &  \textbf{17.38}   \\
    \hline
\end{tabular}
\caption{The GPT-score and Rouge result of different types of memory in learning new knowledge. }
\label{table3}
\end{table}

% msc multi-view table
% \begin{table}[t]
% \centering
% \begin{tabular}{lccc}
%     \hline
%      Memory  &  GPT-score &  Rouge1 & Rouge2 \\
%     \hline
%     his+sum   & 0.57 &  40.58  &  15.48    \\
%     our+his   & 0.58 &  40.73   & 15.52     \\
%     our+sum   & 0.68 &  47.25   & 18.13    \\
%     %our+sum+his   &    &      &      \\
%     \hline
% \end{tabular}
% \caption{The result of learning personal knowledge on multi-view memory.}
% \label{table1}
% \end{table}

% continue+refine multi-view GSB image
% \begin{figure}[t]
% \centering
% \includegraphics[width=1\columnwidth]{fig/compare_multiview.pdf} 
% \caption{The comparison result of learning from feedback and multi-session dependency. Preference for the outputs generated with multi-view memory (red), with only selective and conditional memory (blue), and ties (purple).}
% \label{fig1}
% \end{figure}

We summarize the experiment result as the following conclusions.

\begin{figure}[t]
\centering
\includegraphics[width=1\columnwidth]{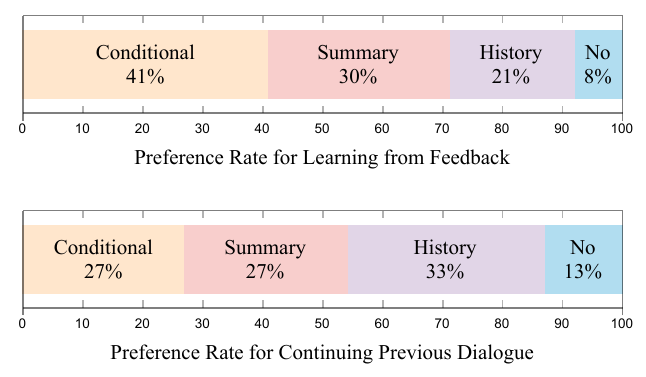} 
\caption{The multi-choice result of different types of memory in learning from feedback and continuing previous dialogue.}
\label{fig5}
\end{figure}

\noindent \textbf{Conditional memory achieves the relatively best result among the three forms of memory.} 
In Table \ref{table3} and Figure \ref{fig5}, we display the effectiveness of different forms of memory on test datasets. The conditional memory achieves the best result in learning new knowledge and learning from human feedback, which shows the ability of conditional memory to abstract and retain important dialogue information. As for continuing previous dialogue, the best result is achieved by directly using dialogue history as memory, and the other two forms of memory also obtain a comparable result. Considering that this task usually requires more details, using dialogue history can have more advantages.

\noindent \textbf{Multi-view memory does not always brings better performance, but the combination between conditional memory and summary-based memory can improve the performance.} 
In Table \ref{table4}, we show the influence of combining conditional memory with the other two memory types as the multi-view memory. On the one hand, we find that history-based memory leads to a performance drop in learning new knowledge and learning from human feedback and a small improvement in continuing previous dialogue. This suggests that there is a overlap between history-based memory and conditional memory and the combination usage of them brings redundancy rather than useful information. On the other hand, we find that summary-based memory is an effective complement to conditional memory and brings improvement in all datasets. 

\begin{table*}[t]
\centering
\begin{tabular}{lccc}
    \hline
     Dataset  &  Conditional &  History & Summary \\
    \hline
    Learning New Knowledge & \textbf{0.65}\ /\ 0.63 &   \textbf{0.59}\ /\ 0.56     & \textbf{0.64}\ /\ 0.57 \\
    Learning from Feedback   & \textbf{28\%}\ /\ 46\%\ /\ 26\%  &   \textbf{33\%}\ /\ 37\%\ /\ 30\%   & \textbf{32\%}\ /\ 40\%\ /\ 28\% \\
    Continuing Previous Dialogue  & \textbf{23\%}\ /\ 57\%\ /\ 20\%  &   26\%\ /\ 44\%\ /\ \textbf{30\%}   & \textbf{33\%}\ /\ 50\%\ /\ 17\% \\
    \hline
\end{tabular}
\caption{The GPT-score and GSB result for the influence of self-reflection retrieval. The GSB test compare the models with and without self-reflection.}
\label{table5}
\end{table*}

\begin{table*}[t]
\centering
\begin{tabular}{lccc}
    \hline
     Dataset  &  No Selection &  No Context &  No Knowledge \\
    \hline
    Learning New Knowledge & \textbf{0.65}\ /\ 0.63     &   0.62\ /\ \textbf{0.63}      &  0.58\ /\ \textbf{0.63} \\
    Learning from Feedback  & 30\%\ /\ 33\%\ /\ \textbf{37\%}  &   27\%\ /\ 30\%\ /\ \textbf{43\%}   &  20\%\ /\ 29\%\ /\ \textbf{51\%} \\
    Continuing Previous Dialogue  & \textbf{30\%}\ /\ 55\%\ /\ 15\%  &   28\%\ /\ 41\%\ /\ \textbf{31\%}   &  21\%\ /\ 52\%\ /\ \textbf{27\%} \\
    \hline
\end{tabular}
\caption{The GPT-score and GSB result for the ablation study. It compares the ablated models and the original ones.}
\label{table6}
\end{table*}

\noindent \textbf{Self-reflection retrieval is effective in most cases, especially on summary-based memory.} 
To evaluate the effectiveness of self-reflection retrieval, we apply it to the LLM assistant with all three types of memory. Here, we limit the self-reflection to no more than one time. As shown in Table \ref{table5}, in most cases, adding a self-reflection retrieval makes the retrieval results more accurate and brings a slight improvement in the outcomes. It is particularly evident in the one with summary-based memory. Considering summary-based memory contains far more information than is required for user input, it leads to relatively poor search accuracy. And the self-reflection retrieval can alleviate this situation.

\begin{figure}[t]
\centering
\includegraphics[width=1\columnwidth]{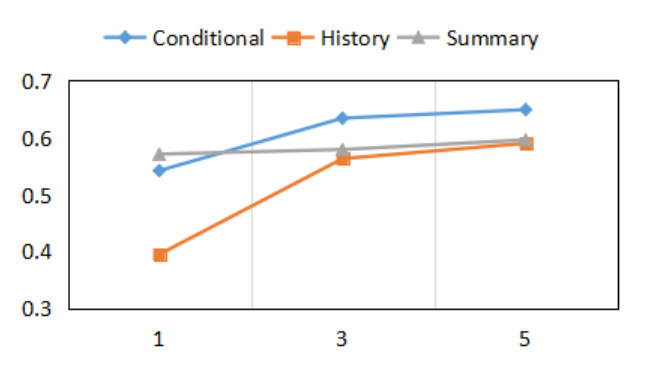} 
\caption{The GPT-score result of different number of retrieved memory records in learning new knowledge.}
\label{fig6}
\end{figure}

\subsection{Analysis and Discussion}

\paragraph{Ablation study on conditional memory}
We further explore how different components contribute to conditional memory. In Table \ref{table6}, we first display the result when we generate and store memory records from all dialogue utterances instead of the selected important ones (No selection). We can find that in situations where crucial information is sparse (learning new knowledge and learning from feedback), utilizing information from all history utterances does not bring significant improvement, and may even result in negative influence. This indicates that conditional memory can filter large amounts of redundancy information. Meanwhile, it is not surprising that more information is better for the task of continuing previous dialogue, when the necessary information may be scattered across the whole history dialogue. But it also brings increasing usage of the LLM in memory construction. We also give the result when we remove the context part (No Context) or knowledge part (No Knowledge) in each conditional memory record. It is clear that the absence of context or knowledge will lead to a decline in the results, which suggests that the information in them is complementary and both are crucial for conditional memory.

\paragraph{Effect of number of retrieved memory records}
In Figure \ref{fig6}, we conduct a brief analysis of the effect of the number of retrieved memory records. For the fine-grained memory records, history-based memory and our conditional memory, the result is no longer significantly improved after $K$ reached 3. Hence, we select the top-3 memory records for history-based memory and conditional memory in our experiment. For the coarse-grained summary-based memory records, the value of $K$ has little effect on the result. Hence, we select one memory record for summary-based memory. Combining with memory records length in Table \ref{tablecase}, adding conditional memory to the assistant increase 140 tokens to the prompt size per utterance on average. This is far more smaller than dialogue history with thousands of tokens, which helps alleviate the problems from having a limited context window.

\section{Conclusion}

In this paper, we propose to use evolving conditional memory to build an easy-access personalized Large Language Model assistant that can learn experience or knowledge during its dialogue with the human user. We focus on the influence of different forms of memory records and the effective usage of the retrieved memory records in generating the response. Moreover, considering the lack of dataset for evaluation in such scenario, we build three test datasets based on different abilities required by a personalized LLM assistant. The experiments show the effectiveness of our proposed methods and bring interesting conclusions.

\section*{Limitations}
One limitation for this work is that we only use GPT-4 rather than other LLM (GPT-3.5, Vicuna) in this paper. And the main reason we use GPT-4 is that it is the strongest LLM now. And we believe this task and the experiments are not easy. So we would like to test the effectiveness of different memory mechanism without the limitation of the LLM. Using weaker model like GPT-3.5 may lead to more results where no model can provide the right answer. But we do think that testing on different LLMs can provide more insights. Meanwhile, it is worth pointing out that we only investigate the foundation of the idea in this paper, and other key points such as the time stamp or forgetting mechanism are still waiting to be explored in further works.

\section*{Ethics Statement}
The scientific artifacts we used are available for research with permissive licenses, including ROUGE and the GPT-4 model. The use of these artifacts is consistent with their intended use.
Considering all the datasets we constructed are built up based on GPT-4 and other existing public available datasets, we think there are no potential ethics risks. All in all, this work complies with the ACL Ethics Policy. 

\bibliography{coling}

\appendix
\label{sec:appendix}

\section{Details for Datasets Construction}

\subsection{Human Evaluation for the Constructed Datasets}
To better understand the quality of the constructed testing dataset, we conduct a small-scale sanity check by human annotators. For each dataset, we randomly select 50 data samples and employ three fluent English speakers to check each of them. The human annotators need to answer two question: (1) Whether the test utterance is related to the source dialogue; (2) Whether the respond in the testing utterance is appropriate based on the source dialogue. If the answers for both questions are yes, the data point is considered as qualified. 

As shown in the Table \ref{table_app1}, we display the qualified percentage of data samples and the Fleiss Kappa coefficient between the three human annotators. We can find that the qualified rate for Learning New Knowledge dataset is relatively lower. The reason is that the dataset is constructed based on multi-session chat (MSC) dataset. In MSC, a set of personal information is firstly given and the annotators are required to conduct the multi-session chat based on these information. In our dataset, we just directly transfer these personal information into question-knowledge pairs for testing. So it is possible that not all personal information are mentioned in the dialogue in MSC dataset. Considering there is over 1000 testing utterance in Learning New Knowledge dataset, these unqualified data sample will not largely effect the experiment result. After all, no model can provide a right answer when facing these testing utterance.

\begin{table}[t]
\centering
\begin{tabular}{lccc}
    \hline
     Dataset  &  Qua  &  Kap  \\
    \hline
    Learning New Knowledge &   60\%  &   0.676       \\
    Learning from Feedback  &  84\%  &   0.746    \\
    Continuing Previous Dialogue  &  81\%   &   0.492    \\
    \hline
\end{tabular}
\caption{Human Evaluation for the Constructed Datasets. Here, ``Qua" stands for the qualified percentage of data samples, and ``Kap" refers to the Fleiss Kappa coefficient between the three human annotators.}
\label{table_app1}
\end{table}

\subsection{Prompt for Continuing Previous Dialogue}

When constructing the test dataset for continuing the previous dialogue, we aim to continue writing a existing dialogue. The related prompt is shown in Table \ref{prompt_continue}.

\begin{table}[t]
    \scriptsize
    \centering
\begin{tabular}{@{}p{\columnwidth}@{}}
\toprule
--------------Input-------------

~

[Dialogue]:

User: .....

Assistant: .....

.....

User: .....

Assistant: .....

~

--------------Prompt-------------

~

The input is a [Dialogue] between a human user and a AI assistant. The background situation is that after the [Dialogue] is finished, some time later, the user want to continue this dialogue with the assistant. Following the previous dialogue, you are reuqired to extend the dialogue to one more turn (user starts the first and then assistant make the respond). The new dialogue turn should follow closely with the [Dialogue]. The new dialogue turn must to utilize or mention the information from the [Dialogue]. In other words, the new dialogue turn can not be independent from the [Dialogue]. The user input can not be respond without the [Dialogue] or the respond is different with or without the [Dialogue]. But do not use pronoun to represent any information in [Dialogue] for the new dialogue. As for the user in the new dialogue, do not apologize, do not compliment, do not thanks, just directly give your question or instruction. The input from the user need to start with "<user>" and the respond from the assistant need to start with "<assistant>".

Example: "<user> XXX <assistant> XXX", where "XXX" refer to the dialogue content.

\\
\bottomrule
\end{tabular}
    \caption{Prompt for constructing continuing previous dialogue dataset.}
    \label{prompt_continue}
\end{table}

\subsection{Prompt for Learning New Knowledge}

For constructing the test dataset for learning new knowledge, we need to transfer a set of personal settings to a question-answering test utterance. Considering it is a relatively easy task, we adopt in-context learning to solve this problem. We display the in-context examples we used in Table \ref{prompt_knowledge}.

\begin{table}[t]
    \scriptsize
    \centering
\begin{tabular}{@{}p{\columnwidth}@{}}
\toprule
--------------Prompt-------------

~

Input: I own a Jeep.

Output: What kind of car do I own? [sep] You own a Jeep.

Input: I enjoy exercising at the gym.

Output: Where do I excercise? [sep] You enjoy exercising at the gym.

Input: I have a marketing job. 

Output: What job do I have? [sep] You have a marketing job.

Input: I don't eat meat.

Output: What kind of food I do not eat? [sep] You don't eat meat.

Input: I am from New England. 

Output: What am I from? [sep] You are from New England.

Input: .....

Output: 

\\
\bottomrule
\end{tabular}
    \caption{Prompt for constructing learning new knowledge dataset.}
    \label{prompt_knowledge}
\end{table}

\subsection{Prompt for Learning from Human Feedback}

To construct the test dataset for learning from human feedback, we decompose it into three steps with different instruction prompts. Given a seed question/instruction, the first step is to generate a similar question/instruction used for testing. Here, we also use in-context learning. The in-context examples are shown in Table \ref{prompt_feedback}. Then we use the LLM to obtain an initial response for the two questions/instructions. The second step is to generate feedback that can be applied to the two both input-response pairs. Hence, we take both pairs as input and use the LLM to generate feedback for them. The instruction prompt is displayed in Table \ref{prompt_feedback2}. With the feedback, we can get an optimized response from the LLM for both seed input and test input. To complete the source dialogue, the last step is to generate a background dialogue for this input-respond-feedback-respond process. The related instruction prompt is shown in Table \ref{prompt_feedback3}. Here, the input in the prompt is the seed question/instruction. 

\begin{table}[t]
    \scriptsize
    \centering
\begin{tabular}{@{}p{\columnwidth}@{}}
\toprule
--------------Prompt-------------

~

Generate a similar question based on the Input. The two question should be the same type and topic but with different details.

Input: Is there anything I can eat for a breakfast that doesn't include eggs, yet includes protein, and has roughly 700-1000 calories? 

Output: Is there anything I can eat for a lunch that doesn't include meat, yet includes protein, and has at least 1000 calories? 

Input: Brainstorm a list of possible New Year's resolutions.

Output: Generate a list of possible Summer Vacation's resolutions. 

Input: Create a fun math question for children.

Output: Create a physics question for children. 

Input: Write a program to compute the sum of integers from k to n.

Output: Write a program to compute the sum of even from 0 to 10.

Input: I am interested in playing Table tennis.

Output: I start to find playing Football is fun.

Input: Let's talk about the famous singer, Taylor Swift. 

Output: Let's talk about Michael Jackson.

Input: .....

Output:

\\
\bottomrule
\end{tabular}
    \caption{Prompt for step 1 of constructing learning from human feedback dataset.}
    \label{prompt_feedback}
\end{table}

\begin{table}[t]
    \scriptsize
    \centering
\begin{tabular}{@{}p{\columnwidth}@{}}
\toprule
--------------Input-------------

~

[Question 1]: .....

[Answer 1]: .....

[Question 2]: .....

[Answer 2]: .....

~

--------------Prompt-------------

~

Given two question-answer pairs, you need to generate one suggestion or critique that improves the quality of the answers for both QA pairs. The suggestion should be generic for both of them and do not focus on very specific content. Only generate the suggestion, do not provide the answers improved by the suggestion.

\\
\bottomrule
\end{tabular}
    \caption{Prompt for step 2 of constructing learning from human feedback dataset.}
    \label{prompt_feedback2}
\end{table}

\begin{table}[t]
    \scriptsize
    \centering
\begin{tabular}{@{}p{\columnwidth}@{}}
\toprule
--------------Input-------------

~

[Input]: .....

~

--------------Prompt-------------

~

You need to generate a multi-turn dialogue between human and assistant in English. The dialogue should end with human saying [Input]. The generated dialogue can be related to the topic of the [Input] or just chatting. 
Example: "<start\_chat> <Human 1>: XXX <Assistant 1>: XXX <Human 2>: XXX <Assistant 2>: XXX <end\_chat>", where "XXX" refer to the actual content of the dialogue. 

The dialogue should follow the following plan:

<start\_chat> 

<Human 1> chat, do not mention [Input] 

<Assistant 1> chat, do not mention [Input]

<Human 2> chat, do not mention [Input]

<Assistant 2> chat, do not mention [Input]

<Human 3> chat, do not mention [Input]

<Assistant 3> chat, do not mention [Input] 

<Human 4> [Input]

<end\_chat>

\\
\bottomrule
\end{tabular}
    \caption{Prompt for step 3 of constructing learning from human feedback dataset.}
    \label{prompt_feedback3}
\end{table}

\section{Details for GPT-4 Evaluation}

Note that the GPT-4 has a position bias and candidates at a specific position (first or second) may have more advantages. Hence, we conduct the evaluation twice and exchange the position of the two candidates. Only when the two evaluation meets the result of each other, the result is definite. If the results are different, we consider it as a tie. Similarly, for multiple choice evaluation, we solve the position bias problem by randomly placing all candidates in different positions. 

In the following section, we introduce the detailed prompts that are used in the evaluation.

\subsection{Prompt for Scoring (GPT-score)}

GPT-score aims to give a predicted response a score between a predefined margin. The LLM is required to give a 0-2 score for a response. Here, ``0” represents the response is totally wrong, ``1” represents the response contains part of personal information in the reference and ``2” represents the response contains all information. We display the related prompt in Table \ref{prompt_scoring}. The input of the prompt is the reference response and the predicted response. 

\begin{table}[t]
    \scriptsize
    \centering
\begin{tabular}{@{}p{\columnwidth}@{}}
\toprule
--------------Input-------------

~

[Reference Answer]: .....

[Predict]: .....

~

--------------Prompt-------------

~

Given a [Reference Answer], is [Predict] covers information in the [Reference Answer]? If the [Predict] covers all the information in the [Reference Answer], generate "2". If the [Predict] covers part of the information in the [Reference Answer], generate "1". If the [Predict] contains no information in the [Reference Answer], generate "0". Do not generate any explanation.

\\
\bottomrule
\end{tabular}
    \caption{Prompt for scoring (GPT-score).}
    \label{prompt_scoring}
\end{table}

\subsection{Prompt for Comparing (GSB test)}

Comparison aims to conduct a Good/Same/Bad test (GSB) between two candidates. The LLM is used to make a decision about which candidate is better given the reference response and other necessary information. As shown in Table \ref{prompt_comparing}, the two candidates are assigned to option A or option B, and we also provide the user input ([Question] in the prompt) and the reference response for the background information.

\begin{table}[t]
    \scriptsize
    \centering
\begin{tabular}{@{}p{\columnwidth}@{}}
\toprule
--------------Input-------------

~

[Question]: .....

[Reference Answer]: .....

A: .....

B: .....

~

--------------Prompt-------------

~

Given a [Question] from user and its corresponding [Reference Answer] from assistant, which candidate answer (A,B) is the best? Best answer refers to the answer that is most similar to the [Reference Answer]. Generate the best answer with "A" or "B", just give the option. Do not give any explanation.

\\
\bottomrule
\end{tabular}
    \caption{Prompt for comparing (GSB test).}
    \label{prompt_comparing}
\end{table}

\subsection{Prompt for Multiple Choice}

Multiple choice is similar to comparison, but the LLM needs to choose the best candidate among A, B, C, and D, four options. The corresponding prompt is shown in Table \ref{prompt_choice}.

\begin{table}[t]
    \scriptsize
    \centering
\begin{tabular}{@{}p{\columnwidth}@{}}
\toprule
--------------Input-------------

~

[Question]: .....

[Reference Answer]: .....

A: .....

B: .....

C: .....

D: .....

~

--------------Prompt-------------

~

Given a [Question] from user and its corresponding [Reference Answer] from assistant, which candidate answer (A,B,C,D) is the best? Best answer refers to the answer that is most similar to the [Reference Answer]. Generate the best answer with "A" or "B" or "C" or "D", just give the option. Do not give any explanation.

\\
\bottomrule
\end{tabular}
    \caption{Prompt for Multiple Choice.}
    \label{prompt_choice}
\end{table}

\section{Details for Methods}

\subsection{Prompt for Summary-Based Memory Construction}
Dialogue summaries can also be used as memory. We use LLM to generate a summary for each dialogue session. The instruction prompt for generating summaries for the dialogue is shown in Table \ref{prompt_summary}.

\begin{table}[t]
    \scriptsize
    \centering
\begin{tabular}{@{}p{\columnwidth}@{}}
\toprule
--------------Input-------------

~

[Dialogue history]:

User: .....

Assistant: .....

.....

User: .....

Assistant: .....

~

--------------Prompt-------------

~

Summarize the [Dialogue history].

\\
\bottomrule
\end{tabular}
    \caption{Prompt for summary-based memory construction.}
    \label{prompt_summary}
\end{table}

\subsection{Prompt for Conditional Memory Construction}
The construction of conditional memory mainly consists of two parts: the decision-making together with the response generation and the memory construction. We display the prompt for the first part in Table \ref{prompt_conditional1}. Here, user input is the input question or instruction from the user and memory stands for the retrieved memory records. It is worth noticing that the context for the current dialogue is not in the prompt, but it is entered as a dictionary like normal LLM assistant. Compared with the normal response generation, we can see that we convert it to a two-step instruction and give an output example for it to better separate the outcome. Only the generated response will be shown to the user and the yes/no decision is hidden behind the model.

We also need to point out that the prompt used here for response generation (only the first step) is also applied to summary-based memory and history-based memory.

\begin{table}[t]
    \scriptsize
    \centering
\begin{tabular}{@{}p{\columnwidth}@{}}
\toprule
--------------Input-------------

~

[User Input]: .....

[Memory]: .....

~

--------------Prompt-------------

~

The first step, generate a response based on the given [User Input] and [Memory]. It should start with "<Respond>: ".

The second step, imagine you are an assistant, make a judgment about the [User Input] on whether it should be remembered. Here are some examples. Answer "yes" if the [User Input] is an requirment for the assistant. Answer "yes" if the [User Input] is a reflection for the assistant previous responds. Answer "yes" if the [User Input] is knowledge worth remembering for the assistant.  Answer "no" if the [User Input] is a simple question that requires the assistant to answer. Answer "no" if the [User Input] is phatic communication like greetings. Only answer "yes" or "no", and do not give any reason. It should start with "<Decision>:" .

Example: "<Respond>: response to the given [User Input] <Decision>: yes or no"

\\
\bottomrule
\end{tabular}
    \caption{Prompt for decision making in conditional memory construction.}
    \label{prompt_conditional1}
\end{table}

For the second part of memory construction, we display its prompt in Table \ref{prompt_conditional2}. Here, the dialogue context refers to ongoing dialogue just including the current utterance. It is also decomposed into three steps. The first step requires the LLM to find the related context from all dialogue contexts. The second step requires the LLM to summarize the related context. And the last step requires the LLM to generate knowledge that is worth memorizing. For the output, we only store the summary and knowledge as the memory record, and the related context is considered to be a mid product.

\begin{table}[t]
    \scriptsize
    \centering
\begin{tabular}{@{}p{\columnwidth}@{}}
\toprule
--------------Input-------------

~

[Dialogue Context]:

User: .....

Assistant: .....

.....

User: .....

Assistant: .....

[User Input]: .....

~

--------------Prompt-------------

~

The first step, copy the context dialogue that related to [User Input] from [Dialogue Context]. Generate "NaN" if there is no dialogue that related to [User Input]. It starts with "<Context>:".

The second step, summarize why the user make [User Input] based on [User Input] itself and the <Context>. It is possible that <Context> is "NaN". If <Context> is "NaN", do not ask for it. The summary needs to describe the situation/reasons/context why the user makes such respond. It starts with "<Summary>:".
The third step, imagine you are an assistant, you would like to better serve the user in the future, rewrite the [User Input] into a note for future reference. The note should also contains the new information/knowledge from [User Input]. It starts with "<Note>:".

Example: "<Context>: dialogue related to current user respond <Summary>: the summary of the dialogue history <Note>: the generated note"

\\
\bottomrule
\end{tabular}
    \caption{Prompt for memory record construction in conditional memory construction.}
    \label{prompt_conditional2}
\end{table}

\subsection{Prompt for Self-Reflection Retrieval}
\begin{table}[t]
    \scriptsize
    \centering
\begin{tabular}{@{}p{\columnwidth}@{}}
\toprule
--------------Input-------------

~

[Candidate memory]:

candidate memory 1

.....

candidate memory n

[User Input]: .....

~

--------------Prompt-------------

~

Imagine you are a helpful assistant and is going to respond to the [User input]. Decide whether the information from the [Candidate memory] is enough to make the respond to the [User input]. If the information is enough, starting with "<Memory>:" and then copy the memory sentences that related to [User input] from [Candidate memory]. If the information is not enough, starting with "<Query>:" and then extend the [User input] with some new keywords to better retrieve the information that you think missing in [Candidate memory].

Example: "<Memory>: the crucial information related to [User input]" or "<Query>: keyword, ..., keyword", "..." refers to unlimited number of keywords.

\\
\bottomrule
\end{tabular}
    \caption{Prompt for self-reflection retrieval.}
    \label{prompt_reflection}
\end{table}

This self-reflection retrieval contains two steps and we show its prompt in Table \ref{prompt_reflection}. The input includes the user input and a set of retrieved candidate memory records. We ask the LLM to make a decision about whether the retrieved information is enough for it to respond to the user input. If it is enough, the LLM assistant selects the related information from these memory records as output. If it is not enough, the LLM assistant is required to improve the original query (user input) to find more information. In practice, the improvement is reified as generating some keywords/phrases representing the missing information.

\end{document}